\documentclass[letterpaper, 10 pt, conference]{ieeeconf}  

\IEEEoverridecommandlockouts                              

\usepackage{amsmath,amssymb,amsfonts}
\usepackage{algpseudocode,algorithm}
\usepackage{subfigure}
\usepackage{graphicx}
\usepackage{cite}

\title{\LARGE \bf
Proximal Policy Optimization with Relative Pearson Divergence*
}

\author{Taisuke Kobayashi$^{1}$
\thanks{*This work was supported by JSPS KAKENHI, Grant-in-Aid for Scientific Research (B), Grant Number JP20H04265.}
\thanks{$^{1}$Taisuke Kobayashi is with the Division of Information Science, Nara Institute of Science and Technology, 8916-5 Takayama-cho, Ikoma, Nara 630-0192, Japan
        {\tt\small kobayashi@is.naist.jp}}%
}

\begin{document}

\maketitle
\thispagestyle{empty}
\pagestyle{empty}

\begin{abstract}

The recent remarkable progress of deep reinforcement learning (DRL) stands on regularization of policy for stable and efficient learning.
A popular method, named proximal policy optimization (PPO), has been introduced for this purpose.
PPO clips density ratio of the latest and baseline policies with a threshold, while its minimization target is unclear.
As another problem of PPO, the symmetric threshold is given numerically while the density ratio itself is in asymmetric domain, thereby causing unbalanced regularization of the policy.
This paper therefore proposes a new variant of PPO by considering a regularization problem of relative Pearson (RPE) divergence, so-called PPO-RPE.
This regularization yields the clear minimization target, which constrains the latest policy to the baseline one.
Through its analysis, the intuitive threshold-based design consistent with the asymmetry of the threshold and the domain of density ratio can be derived.
Through four benchmark tasks, PPO-RPE performed as well as or better than the conventional methods in terms of the task performance by the learned policy.

\end{abstract}


\section{Introduction}

The next-generation robots aim to resolve complicated tasks with unknown dynamics: e.g. human assistance~\cite{modares2015optimized} and cloth manipulation~\cite{tsurumine2019deep}.
Reinforcement learning (RL)~\cite{sutton2018reinforcement} is the promising methodology for them due to model-free approach.
Indeed, its extension using deep neural networks (DNNs)~\cite{lecun2015deep} to approximate policy and value functions, named deep RL (DRL)~\cite{silver2016mastering}, has received a lot of attention.


The recent remarkable progress of DRL stands on policy regularization.
By constraining the update amount and direction of policy, the policy regularization allows the policy to improve smoothly toward the optimal one.
Several variations of regularization way have been proposed:
e.g. hard/soft constraint by Kullback-Leibler (KL) divergence between the latest and baseline policies~\cite{schulman2015trust,tsurumine2019deep}; maximization of entropy of policy~\cite{haarnoja2018soft}; indirect minimization of variance of temporal difference (TD)~\cite{parisi2019td}; and so on.
All of them contribute significantly to improving learning performance and are treated as representatives of DRL.
Note that these recent studies have been systematized as policy-regularized RL~\cite{geist2019theory}.

Among the methods for policy regularization, this paper further focuses on proximal policy optimization (PPO)~\cite{schulman2017proximal}, which has been proposed as soft constraint version of KL divergence.
Therefore, the first implementation of PPO is with a direct regularization of KL divergence between the latest and baseline policies.
The gain of this regularization should be adjusted for facing situations, but the proposed way was in an ad-hoc manner, which could not fully optimize the gain.
For more intuitive and practical implementation, another version, which is more popular than the first one, assumes the use of importance sampling and clips density ratio of the latest and baseline policies with a threshold.
By doing so, the policy gradient vanishes when the density ratio is clipped, and that constrains the update of the policy.
From here, PPO refers to this clipping version.

However, PPO has two theoretical problems at the cost of its intuitive implementation.
First one comes from the modification from the soft constraint of KL divergence to clipping the density ratio, namely, its minimization target is unclear.
Indeed, the follow-up study of PPO~\cite{wang2020truly} analyzed its behavior and pointed out that PPO has no capability to make the latest policy bind to the baseline one.
As another problem, the symmetric threshold is given numerically while the density ratio itself is in asymmetric domain.
This gap may induce unbalanced regularization of the policy.

Hence, this paper tackles these two theoretical issues in PPO by considering a new regularization problem of relative Pearson (RPE) divergence~\cite{yamada2013relative,sugiyama2013direct}.
This RPE-divergence-based regularization yields the clear minimization target, which can softly constrain the latest policy to the baseline one.
The symmetric density ratio can be obtained by adjusting a relativity parameter.
To inherit the intuitive threshold-based design of PPO, the gain of RPE regularization is converted to the corresponding threshold through mathematical derivation.
The proposed method, so-called PPO-RPE, therefore achieves both the intuitive implementation by the threshold and solving the theoretical issues in PPO.

To investigate the effects of solving the theoretical issues in PPO by PPO-RPE, four benchmark tasks are simulated in a Pybullet simulator~\cite{coumans2016pybullet}.
With comparison to the conventional methods (i.e. PPO and PPO-RB~\cite{wang2020truly}), PPO-RPE achieved the equivalent/superior task performance in all the tasks.
During training the tasks, it is found that PPO-RPE regularizes the policy more than the conventional methods to explicitly minimize the RPE divergence between the latest and baseline policies.
As a consequence, it can be indicated that PPO-RPE contributes to improvement of the learning performance by restricting the update of the latest policy in a theoretically rigorous way.

\section{Preliminaries}

\subsection{Policy-regularized reinforcement learning}

The original purpose of RL~\cite{sutton2018reinforcement} is that an agent optimizes its policy $\pi$ in order to maximize the sum of rewards $r$ (i.e. a return $\sum_{k=0}^\infty \gamma^k r_{k}$ with $\gamma \in [0, 1)$ discount factor) received from an environment.
To this end, numerous interactions, which exchange actions $a$ sampled from the agent with states $s$ sampled from the environment according to Markovian dynamics, are performed during learning $\pi$.
This paper considers the minimization problem of the expected negative return at $t$ time step, in other words the expected negative action-value function $Q(s_t, a_t)$, as follows:
\begin{align}
    \pi^*(a_t \mid s_t) &= \arg \min_{\pi} - \mathbb{E}_{a_t \sim \pi} \left [ A(s_t, a_t) \right ]
    \label{eq:piopt_std}
\end{align}
where $A(s_t, a_t) = Q(s_t, a_t) - V(s_t)$ with $V(s_t)$ denotes the advantage function with the state-value function.
$V(s_t)$ can be added to the minimization target since its gradient w.r.t policy is equal to zero.
The optimal policy $\pi^*$ is obtained through this minimization problem.

Recently, to improve RL performance in terms of fast convergence and reducing variance of results, regularization terms for the policy $\Omega$ is additionally minimized~\cite{geist2019theory}.
That is, the following surrogated minimization target can unify most of the recent RL algorithms (e.g.~\cite{schulman2017proximal,haarnoja2018soft,parisi2019td}).
\begin{align}
    \mathcal{L}^\dagger(\pi) &= - \mathbb{E}_{a_t \sim \pi} \left [ A(s_t, a_t) \right ] + \mathbb{E}_{a_t \sim \pi} \left [ \Omega(s_t, a_t) \right ]
    \nonumber \\
    &= - \mathbb{E}_{a_t \sim \pi_b} \left [ \frac{\pi(a_t \mid s_t)}{\pi_b(a_t \mid s_t)} \left ( A(s_t, a_t) - \Omega(s_t, a_t) \right ) \right ]
    \nonumber \\
    &= - \mathbb{E}_{a_t \sim \pi_b} \left [ \rho(s_t, a_t) A^\dagger(s_t, a_t) \right ]
    \label{eq:loss_reg}
\end{align}
where by introducing a baseline policy $\pi_b(a_t \mid s_t)$ (e.g. the old version of $\pi$ or the one outputted from slowly updated target network~\cite{mnih2015human}) for importance sampling, a density ratio $\rho = \pi / \pi_b$ is derived in the minimization target, which is clipped in PPO~\cite{schulman2017proximal,wang2020truly} (details are in the next section).
The optimal policy in the policy-regularized optimization problem, $\pi^\dagger$, is obtained by minimizing $\mathcal{L}^\dagger$.

Here, a policy-gradient method is generally employed for solving this optimization problem.
That is, when the policy is parameterized by a parameter set $\theta$ (e.g. weights and biases in deep neural networks), $\theta$ is updated by its first-order gradient as follows:
\begin{align}
    \nabla_\theta \mathcal{L}^\dagger(\pi) &= - \mathbb{E}_{a_t \sim \pi_b} \left [ \nabla_\theta \{ \rho(s_t, a_t) A^\dagger(s_t, a_t) \} \right ]
    \nonumber \\
    &\simeq - \rho(s_t, a_t) \tilde{A}^\dagger(s_t, a_t) \nabla_\theta \ln \pi(a_t \mid s_t)
    \label{eq:grad_reg} \\
    \theta &\gets \theta - \alpha \mathrm{SGD}(\nabla_\theta \mathcal{L}^\dagger(\pi))
    \label{eq:sgd}
\end{align}
where $\tilde{A}^\dagger = A^\dagger + \pi \nabla_\pi A^\dagger = A^\dagger + \rho \nabla_\rho A^\dagger$.
If $A^\dagger$ has no direct computational graph with $\pi$, $\tilde{A}^\dagger = A^\dagger$.
Monte-Calro approximation, where only one action is sampled from the baseline policy, is applied to omit the use of closed-form solution of the expectation.
In general DRL, $\theta$ is updated using one of the stochastic gradient descent (SGD) optimizers like~\cite{ilboudo2020robust}.

\subsection{Proximal policy optimization}

In PPO~\cite{schulman2017proximal}, the initial version had the explicit regularization for KL divergence.
However, its adaptive gain could not be designed properly, and this version did not achieve significant performance.
To improve the performance with the practical implementation, the new version of PPO and its variant (named PPO-RB~\cite{wang2020truly}) therefore give eq.~\eqref{eq:loss_reg} the following condition according to a threshold parameter $\epsilon$.
That is, the following surrogated density ratio $\rho^\mathrm{PPO}$ replaces the original $\rho$ in eq.~\eqref{eq:loss_reg}, instead of explicit definition of $\Omega$.
\begin{align}
    \rho^\mathrm{PPO} &= \begin{cases}
        -\eta \rho + (1 + \eta)(1 + \sigma\epsilon) & \sigma (\rho - 1) \geq \epsilon
        \\
        \rho & \mathrm{otherwise}
    \end{cases}
    \label{eq:ratio_ppo} \\
    \mathcal{L}^\mathrm{PPO}(\pi) &= - \mathbb{E}_{a_t \sim \pi_b} \left [ \rho^\mathrm{PPO}(s_t, a_t) A(s_t, a_t) \right ]
    \label{eq:loss_ppo}
\end{align}
where $\sigma$ denotes the sign of $A$, and $\eta \geq 0$ is for rollback to the baseline policy.
If $\eta = 0$, this problem is equivalent to the original PPO.

With this condition for clipping if $\eta = 0$ or rollback if $\eta > 0$, PPO inhibits update of the latest policy $\pi$ away from the baseline $\pi_b$, although its specific regularization target represented by $\Omega$ is complicated.
Note that it is reported that only the case with $\eta > 0$ enables $\pi$ to bind with $\pi_b$~\cite{wang2020truly}.

\subsection{Relative Pearson divergence}

\begin{figure}[tb]
    \centering
    \subfigure[3D view]{
        \includegraphics[keepaspectratio=true,width=0.45\linewidth]{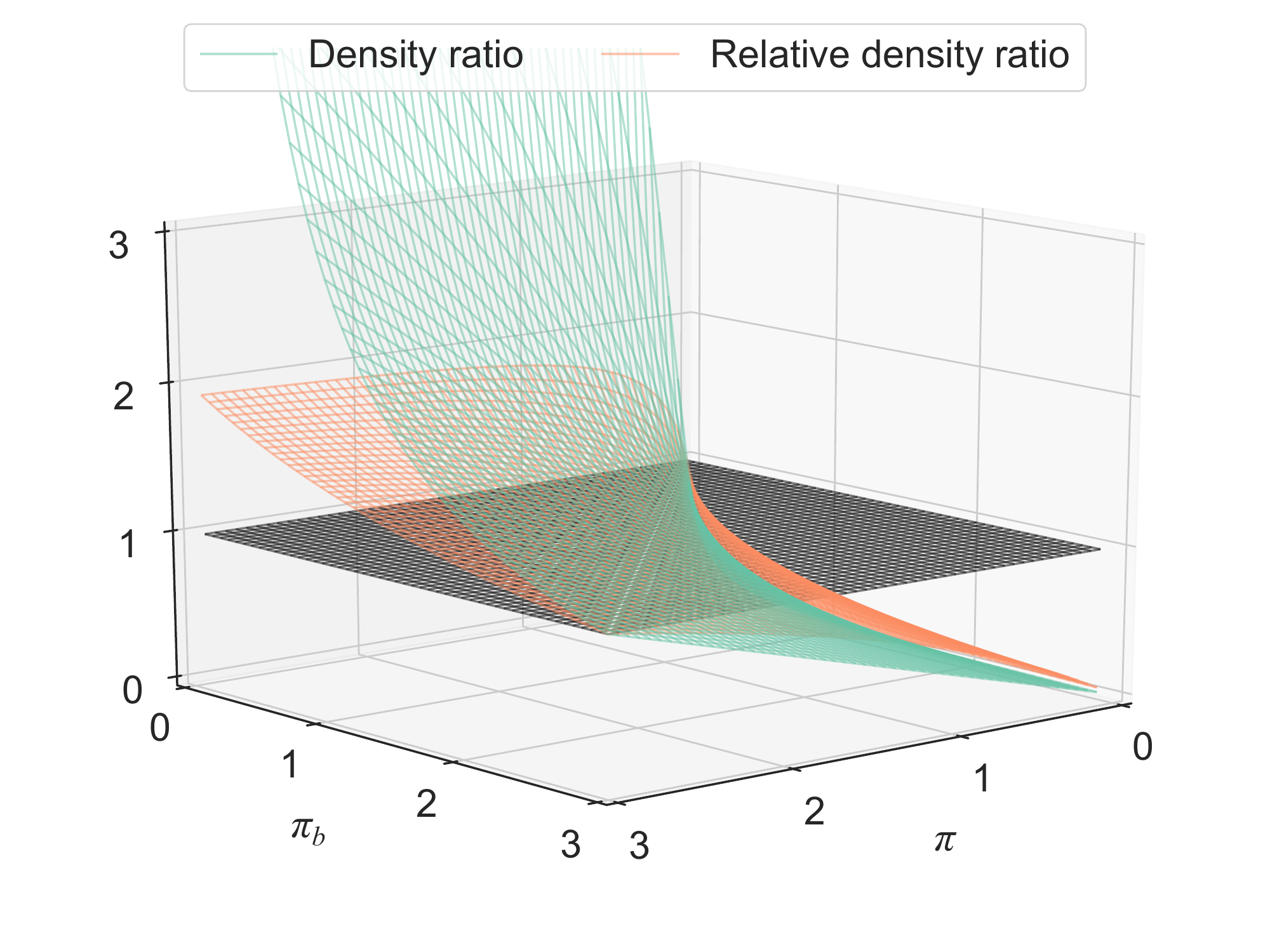}
    }
    \centering
    \subfigure[2D view]{
        \includegraphics[keepaspectratio=true,width=0.45\linewidth]{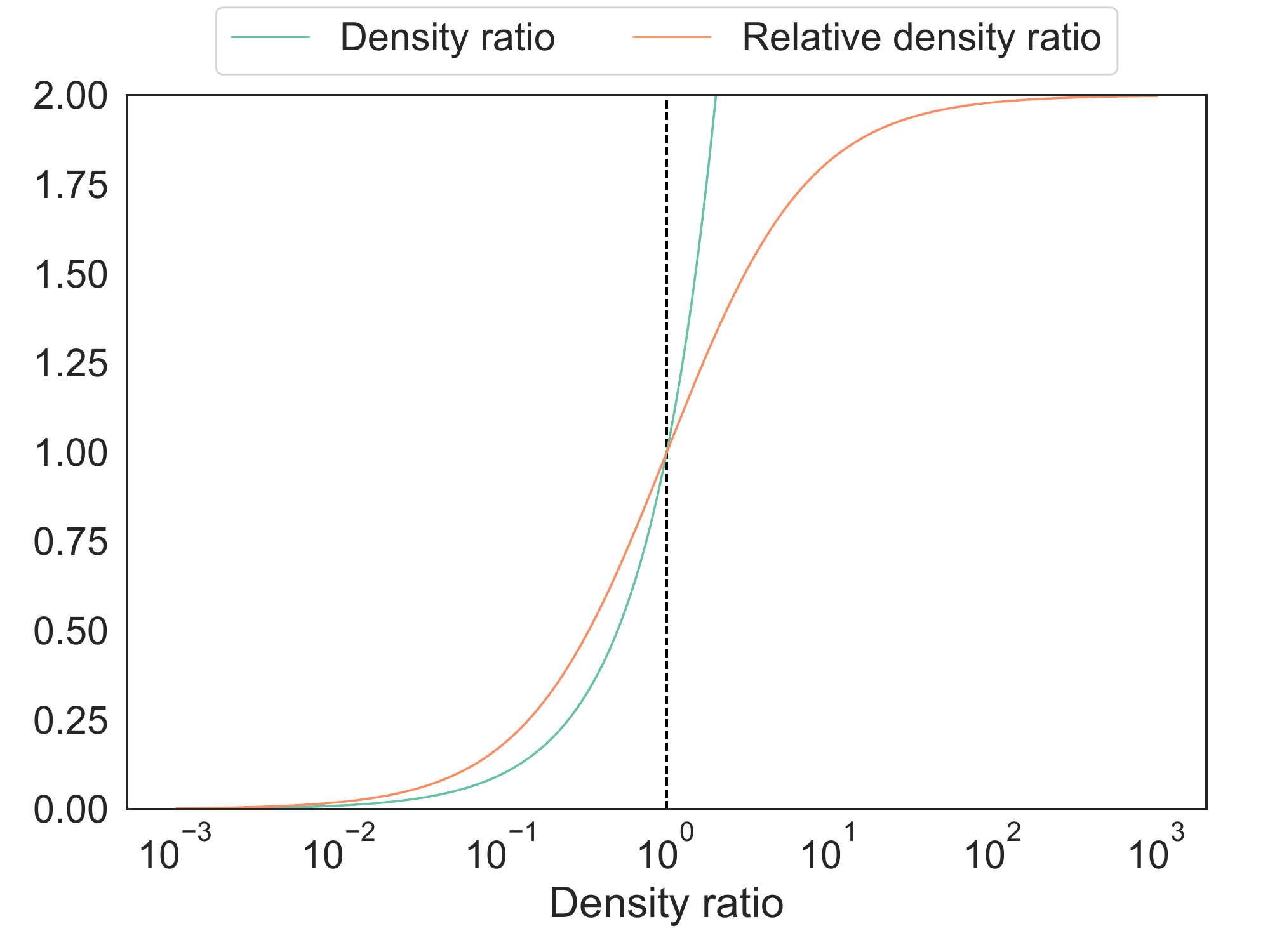}
    }
    \caption{Relative density ratio with $\beta = 0.5$:
        if and only if $\pi = \pi_b$, the both density ratios are equal to one;
        although the raw density ratio diverges as $\pi_b$ decreases and $\pi$ increases (i.e. $\rho$ increases), the relative density ratio has finite upper bounds $1/\beta$;
        in particular, if $\beta = 0.5$, the shape of relative density ratio becomes symmetric around one as shown in (b).
    }
    \label{fig:img_symmetric}
\end{figure}

To measure the divergence between two probabilities, KL divergence is the most major metric.
However, it is actually the special case of f-divergence, which is the generalized divergence~\cite{liese2006divergences}, and therefore, we can measure the divergence using either of other f-divergences.
To inherit the behaviors of PPO (eqs.~\eqref{eq:ratio_ppo} and~\eqref{eq:loss_ppo}), the following conditions are required for selecting the specific f-divergence.
\begin{enumerate}
    \item When $\rho = 1$, the term inside the expectation of the loss function should be $\rho A$.
    \item Only when $\rho$ is on the threshold, the gradient must be zero and there is only one vertex.
\end{enumerate}
In addition, the symmetry is also desired.
Unfortunately, it is found that KL divergence cannot satisfy all of them simultaneously.
Thus, the alternative divergence that satisfies all the above is heuristically given as follows.

Specifically, Pearson (PE) divergence, which is one of the f-divergences, is utilized while modifying it to the relative version, RPE divergence~\cite{yamada2013relative,sugiyama2013direct}.
The raw PE divergence between two policies, $\pi$ and $\pi_b$, is given as follows:
\begin{align}
    \mathrm{PE}(\pi, \pi_b) = \mathbb{E}_{a \sim \pi_b} \left [ \frac{1}{2}\left ( \rho(a) - 1 \right )^2 \right ]
    \label{eq:div_pe}
\end{align}
Since the mean of $\rho$ is always equal to $1$ (i.e. $\int \pi_b(a \mid s) \rho(s, a) da = \int \pi(a \mid s) da = 1$), this divergence computes the mean squared error of $\rho$ from its mean.
This divergence is non-negative and vanishes if and only if $\pi = \pi_b$.

When the denominator $\pi_b$ has small value, the numerical computation of PE divergence would diverge.
For numerical stability, RPE divergence first introduces the following mixture probability with a mixture ratio $\beta \in [0, 1]$.
\begin{align}
    \pi_\beta(a \mid s) = \beta \pi(a \mid s) + (1 - \beta) \pi_b(a \mid s)
    \label{eq:pi_beta}
\end{align}
Using this, RPE divergence is defined as follows:
\begin{align}
    \mathrm{PE}_\beta(\pi, \pi_b) = \mathbb{E}_{a \sim \pi_\beta} \left [ \frac{1}{2}\left ( \frac{\rho(a)}{\beta \rho(a) + (1 - \beta)} - 1 \right )^2 \right ]
    \label{eq:div_rpe}
\end{align}
where the relative density ratio between $\pi$ and $\pi_\beta$, $\rho_\beta = \pi / \pi_\beta$, is within $[0, 1/\beta)$.
The numerical stability is, therefore, guaranteed as expected, while KL divergence is numerically unstable due to the use of logarithm and raw density ratio.
In particular, when $\beta = 0.5$, the range of the density ratio is bounded in $[0, 2)$ while it has the mean on $1$, that is, it has a symmetric distribution around one (see Fig.~\ref{fig:img_symmetric}).
This symmetry of the relative density ratio yields the asymmetric thresholds in the proposed method.

\section{Proximal policy optimization with relative Pearson divergence: PPO-RPE}

\subsection{Overview}

\begin{algorithm}[tb]
    \caption{PPO-RPE}
    \label{alg:pporpe}
    \begin{algorithmic}[1]
        \State{Set $\epsilon \in (0, 1)$ (0.1--0.3 is the recommended range)}
        \State{Set $\beta \in [0, 1]$ (0.5 is the default value)}
        \State{Initialize the policy $\pi$ with $\theta$ and the baseline policy $\pi_b$}
        \State{Initialize optimizer as SGD with learning rate $\alpha$}
        \While{True}
            \State{Get $s_t$ from the environment}
            \State{Sample $a_t$ from the baseline policy $\pi_b(a_t \mid s_t)$}
            \State{Get $s_{t+1}, r_t$ from the environment by acting on $a_t$}
            \State{$A = Q(s_t, a_t) - V(s_t)$}
            \State{$\sigma = \mathrm{sign}(A)$}
            \State{$\rho = \cfrac{\pi(a_t \mid s_t)}{\pi_b(a_t \mid s_t)}$}
            \State{$\rho_\beta = \cfrac{\rho}{1 - \beta + \beta \rho}$}
            \State{$\mathrm{denom} = \sigma \epsilon \left [\beta \sigma \epsilon - 2 \left \{1 - \beta (1 + \sigma \epsilon) \right \} \right ]$}
            \State{$\mathrm{numer} = A (\rho_\beta - 1) \left \{\beta (\rho_\beta - 1) - \cfrac{2 (1 - \beta)}{1 - \beta + \beta \rho} \right \}$}
            \State{$\tilde{A}^\mathrm{RPE} = A - \cfrac{\mathrm{numer}}{\mathrm{denom}}$}
            \State{$g = - \rho \tilde{A}^\mathrm{RPE} \nabla_\theta \ln \pi(a_t \mid s_t)$}
            \State{$\theta \gets \theta - \alpha \mathrm{SGD}(g)$}
            \State{Update $\pi_b$ according to $\pi$ with $\theta$}
            \State{$t \gets t + 1$}
        \EndWhile
    \end{algorithmic}
\end{algorithm}

\begin{figure}[tb]
    \centering
    \includegraphics[keepaspectratio=true,width=0.98\linewidth]{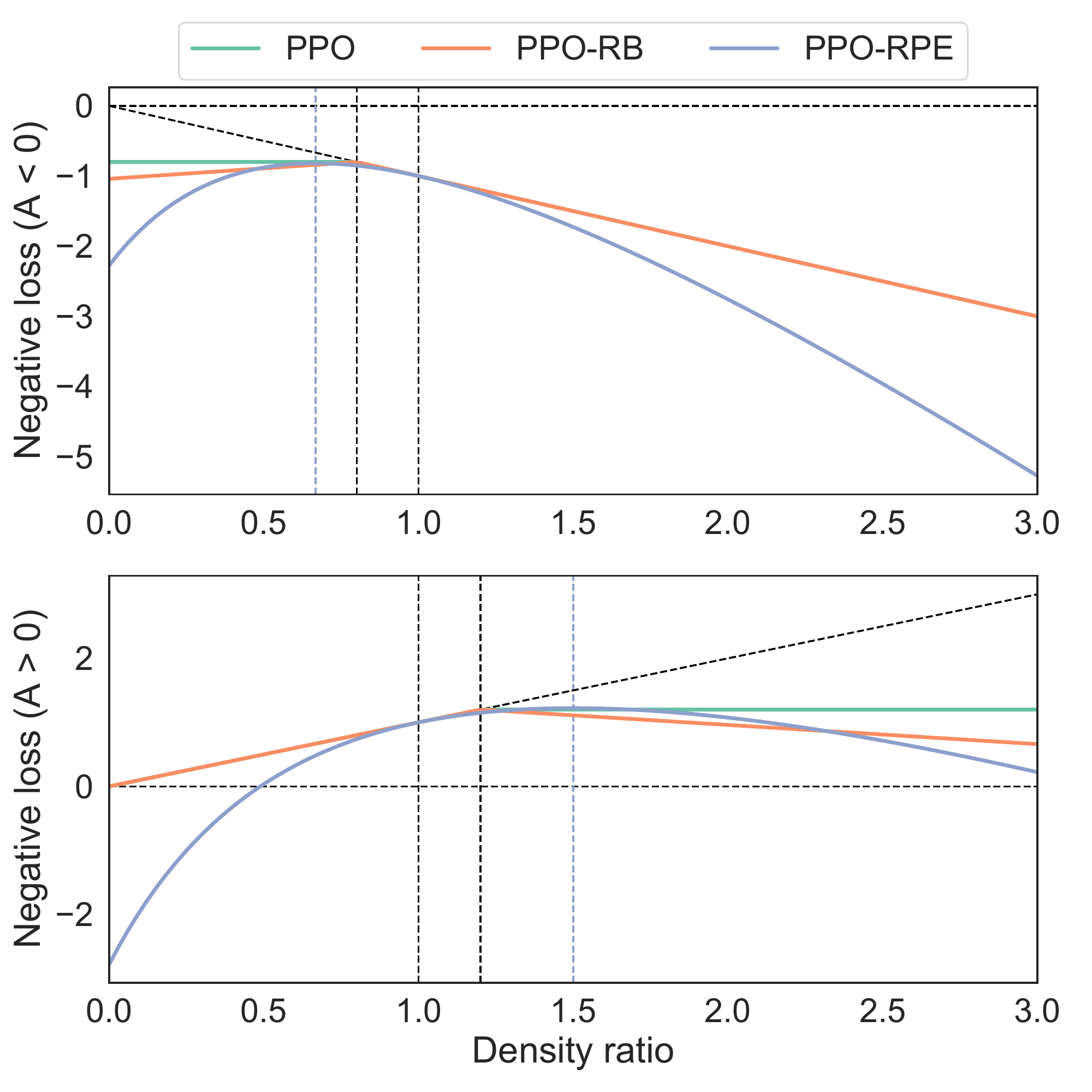}
    \caption{Negative loss functions for the respective methods:
        PPO and PPO-RB place the two thresholds (when $A > 0$ and $A < 0$, respectively) symmetrically around one;
        in contrast, PPO-RPE sets its two vertices (when $A > 0$ and $A < 0$, respectively) asymmetrically in the domain of the raw density ratio;
        in addition, PPO-RPE has the capability to bind the latest policy to the baseline policy.
    }
    \label{fig:img_loss}
\end{figure}

The issues in PPO and PPO-RB are listed below:
\begin{itemize}
    \item No explicit regularization term is given, thereby making the problem mathematically difficult to understand what divergence policy is regularized in terms of.
    \item No capability to bind the latest policy to the baseline policy has been reported in~\cite{wang2020truly}, where ad-hoc implementations have been proposed to solve it.
    \item Inconsistency between symmetric design of the threshold and asymmetric domain of the density ratio would induce unbalanced regularization.
\end{itemize}

To resolve them by a mathematically rigorous way, this paper proposes a proximal policy optimization with relative Pearson divergence, so-called PPO-RPE.
Specifically, PPO-RPE defines the minimization problem with RPE divergence regularization.
Since the regularization target is mathematically given, the capability to bind the latest policy to the baseline one can easily be guaranteed.
In addition, the gain of its regularization is adaptively designed based on a symmetric threshold for the relative density ratio in RPE divergence, which makes the threshold for the raw density ratio asymmetric.

Here, the overall implementation of PPO-RPE is summarized in Alg.~\ref{alg:pporpe}, while the details are described from the next sections.
In addition, all the examples of the negative loss functions (i.e. $\rho A^\dagger$) for the proposed and conventional methods are illustrated in Fig.~\ref{fig:img_loss}.
As can be seen in this figure, the previous methods (i.e. PPO and PPO-RB) regularize only when $\rho$ exceeds one side ($\rho > 1 + \epsilon$ when $A > 0$ and $\rho < 1 - \epsilon$ when $A < 0$).
In that case, the latest policy $\pi$ can leave from the baseline one $\pi_b$ without constraint to $\pi_b$ for example when $\rho < 1$ and $A > 0$.
In contrast, PPO-RPE has the capability to bind $\pi$ to $\pi_b$ in accordance with the asymmetric threshold for $\rho$.

\subsection{Minimization target}

As a first step to explicitly regard PPO-RPE as a sub-class of policy-regularized RL described in eq.~\eqref{eq:loss_reg}, the regularization term for PPO-RPE, $\Omega^\mathrm{RPE}$, is defined.
Note that the arguments $(s_t, a_t)$ in $A$, $\rho$, and $\rho_\beta$ are omitted from here for brevity.

With $\pi_\beta$ in eq.~\eqref{eq:pi_beta} and $\rho_\beta = \pi / \pi_\beta = \rho / (1 - \beta + \beta \rho)$, RPE-divergence-based $\Omega^\mathrm{RPE}$ is given as follows:
\begin{align}
    \Omega^\mathrm{RPE} &= C \frac{(\rho_\beta - 1)^2}{\rho_\beta}
    = C \frac{1 - \beta + \beta \rho}{\rho} (\rho_\beta - 1)^2
    \label{eq:reg_rpe}
\end{align}
where $C$ denotes the gain of this regularization.
The expectation w.r.t $\pi$ of this regularization is equivalent to eq.~\eqref{eq:div_rpe} amplified by $C$.

Finally, by substituting eq.~\eqref{eq:reg_rpe} for eq.~\eqref{eq:loss_reg}, we have the following loss function to be minimized for optimization of the policy.
\begin{align}
    \mathcal{L}^\mathrm{RPE}(\pi) &= - \mathbb{E}_{a_t \sim \pi_b} \left [ \rho A - C (1 - \beta + \beta \rho) (\rho_\beta - 1)^2 \right ]
    \nonumber \\
    &= - \mathbb{E}_{a_t \sim \pi_b} \left [ \rho A^\mathrm{RPE} \right ]
    \label{eq:loss_rpe}
\end{align}
where the regularization term is included in $A^\mathrm{RPE}$.
Note that $\rho A^\mathrm{RPE} / A$ can be regarded as $\rho^\mathrm{RPE}$ corresponding to PPO.

\subsection{Policy gradient}

To analyze the behavior of PPO-RPE and to design $C$ (see the next section), the policy gradient is analytically derived.
Since we know $\nabla_\theta \rho = \rho \nabla_\theta \ln \pi$ and the expectation of the loss function is approximated by Monte Carlo method, what we have to derive is the gradient of $\mathcal{L}^\mathrm{RPE}$ w.r.t the density ratio, which is equivalent to the negative surrogated advantage function: $\nabla_\rho \mathcal{L}^\mathrm{RPE} \simeq - A^\mathrm{RPE} - \rho \nabla_\rho A^\mathrm{RPE} = - \tilde{A}^\mathrm{RPE}$.
Now, $\tilde{A}^\mathrm{RPE}$ can be computed analytically as follows:
\begin{align}
    \tilde{A}^\mathrm{RPE} &\simeq A - C \beta (\rho_\beta - 1)^2 - 2 C (1 - \beta + \beta \rho) (\rho_\beta - 1) \frac{\partial \rho_\beta}{\partial \rho}
    \nonumber \\
    &= A \left [ 1 - \frac{C}{A}(\rho_\beta - 1)\left \{ \beta (\rho_\beta  - 1) + \frac{2 (1 - \beta)}{1 - \beta + \beta \rho} \right \} \right ]
    \label{eq:grad_rpe}
\end{align}
where $\partial \rho_\beta / \partial \rho = (1 - \beta) / (1 - \beta + \beta \rho)^2$.
This is substituted for eq.~\eqref{eq:grad_reg}, and used for updating the parameter set $\theta$ in eq.~\eqref{eq:sgd}.

The second term in the square bracket is for the regularization, and vanishes if and only if $\rho = \rho_\beta = 1$ (i.e. $\pi = \pi_b$) regardless of $C$ and $\beta$.
Therefore, we can see that, with this policy gradient, the policy $\pi$ is appropriately updated toward maximization of the advantage function $A$ (i.e. the action-value function $Q$) while being regularized to the baseline policy $\pi_b$.

\subsection{Threshold-based design of gain for regularization}

The optimal gain of the regularization $C$ is difficult to be designed due to its non-intuitiveness.
In addition, the constant $C$ is insufficient to adapt the agent to various environments without tuning $C$ for each environment.
Therefore, inspired by PPO, the threshold-based design of $C$ is derived.

First of all, we focus on the fact that the density ratio $\rho$ is asymmetric around one, and therefore, the symmetric threshold for it as conducted in PPO would cause unbalanced regularization.
Instead, PPO-RPE sets the threshold for the relative density ratio $\rho_\beta$, which can yield the symmetric domain if $\beta = 0.5$ (see Fig.~\ref{fig:img_symmetric}).
Given the threshold parameter $\epsilon \in (0, 1)$, the following condition is considered.
\begin{align}
    \rho_\beta^\epsilon &= 1 + \sigma \epsilon
    \label{eq:thr_rb} \\
    \rho^\epsilon &= \frac{(1 - \beta) (1 + \sigma \epsilon)}{1 - \beta (1 + \sigma \epsilon)}
    = 1 + \frac{\sigma \epsilon}{1 - \beta (1 + \sigma \epsilon)}
    \label{eq:thr_r}
\end{align}
where $\rho^\epsilon$ can be easily derived using the fact $\rho_\beta = \rho / (1 - \beta + \beta \rho)$.
Note that the term $\sigma \epsilon / (1 - \beta (1 + \sigma \epsilon))$ in $\rho^\epsilon$ yields the asymmetric threshold for $\rho$, and if $\beta = 0$ with no relativity, it is reverted to be symmetric.

At $\rho_\beta^\epsilon$ (and $\rho^\epsilon$), to bind the latest policy $\pi$ to the baseline $\pi_b$, $\tilde{A}^\mathrm{RPE}$ is desired to be zero.
That is, $C$ can be derived from the terms in the square bracket of eq.~\eqref{eq:grad_rpe} as follows:
\begin{align}
    0 &= 1 - \frac{C}{A} (\rho_\beta^\epsilon - 1) \left \{ \beta (\rho_\beta^\epsilon  - 1) + \frac{2 (1 - \beta)}{1 - \beta + \beta \rho^\epsilon} \right \}
    \nonumber \\
    &= 1 - \frac{C}{A} \sigma \epsilon \left \{ \beta \sigma \epsilon + 2 (1 - \beta) \frac{1 - \beta (1 + \sigma \epsilon)}{1 - \beta} \right \}
    \nonumber \\
    \frac{C}{A} &= \frac{1}{\sigma \epsilon [ \beta \sigma \epsilon + 2 \{1 - \beta (1 + \sigma \epsilon)\} ]}
    \label{eq:mag_rpe}
\end{align}
This equation shows that $C$ depends on the scale of $A$, namely, it has adaptability to various environments with different scales of reward function.

By substituting this for eq.~\eqref{eq:grad_rpe}, PPO-RPE achieves RPE-divergence-based regularization with the threshold-based gain.
\begin{align}
    \tilde{A}^\mathrm{RPE} = A - \cfrac{A (\rho_\beta - 1) \left \{\beta (\rho_\beta - 1) - \cfrac{2 (1 - \beta)}{1 - \beta + \beta \rho} \right \}}{\sigma \epsilon \left [\beta \sigma \epsilon - 2 \left \{1 - \beta (1 + \sigma \epsilon) \right \} \right ]}
    \label{eq:grad_rpe_thre}
\end{align}
Finally, the parameter set $\theta$ is updated according to SGD with the policy gradient $- \rho \tilde{A}^\mathrm{RPE} \nabla_\theta \ln \pi(a_t \mid s_t)$.

\section{Simulations}



\begin{table}[tb]
    \caption{Common hyperparameters for the simulations}
    \label{tab:sim_parameter}
    \centering
    \begin{tabular}{ccc}
        \hline\hline
        Symbol & Meaning & Value \\
        \hline
        $\gamma$ & Discount factor & 0.99 \\
        $\alpha$ & Learning rate & 3e-4 \\
        $(\tau, \nu)$ & Hyperparameters for~\cite{kobayashi2021t} & (0.3, 1.0) \\
        $(\lambda_\mathrm{max}^1, \lambda_\mathrm{max}^2, \kappa)$ & Hyperaparameters for~\cite{kobayashi2020adaptive} & (0.5, 0.95, 10) \\
        $\epsilon$ & Threshold parameter & 0.1 \\
        $\beta_{DE}$ & Gain for entropy regularization & 0.01 \\
        $\beta_{TD}$ & Gain for TD regularization & 0.01 \\
        \hline\hline
    \end{tabular}
\end{table}

\begin{figure}[tb]
    \centering
    \includegraphics[keepaspectratio=true,width=0.98\linewidth]{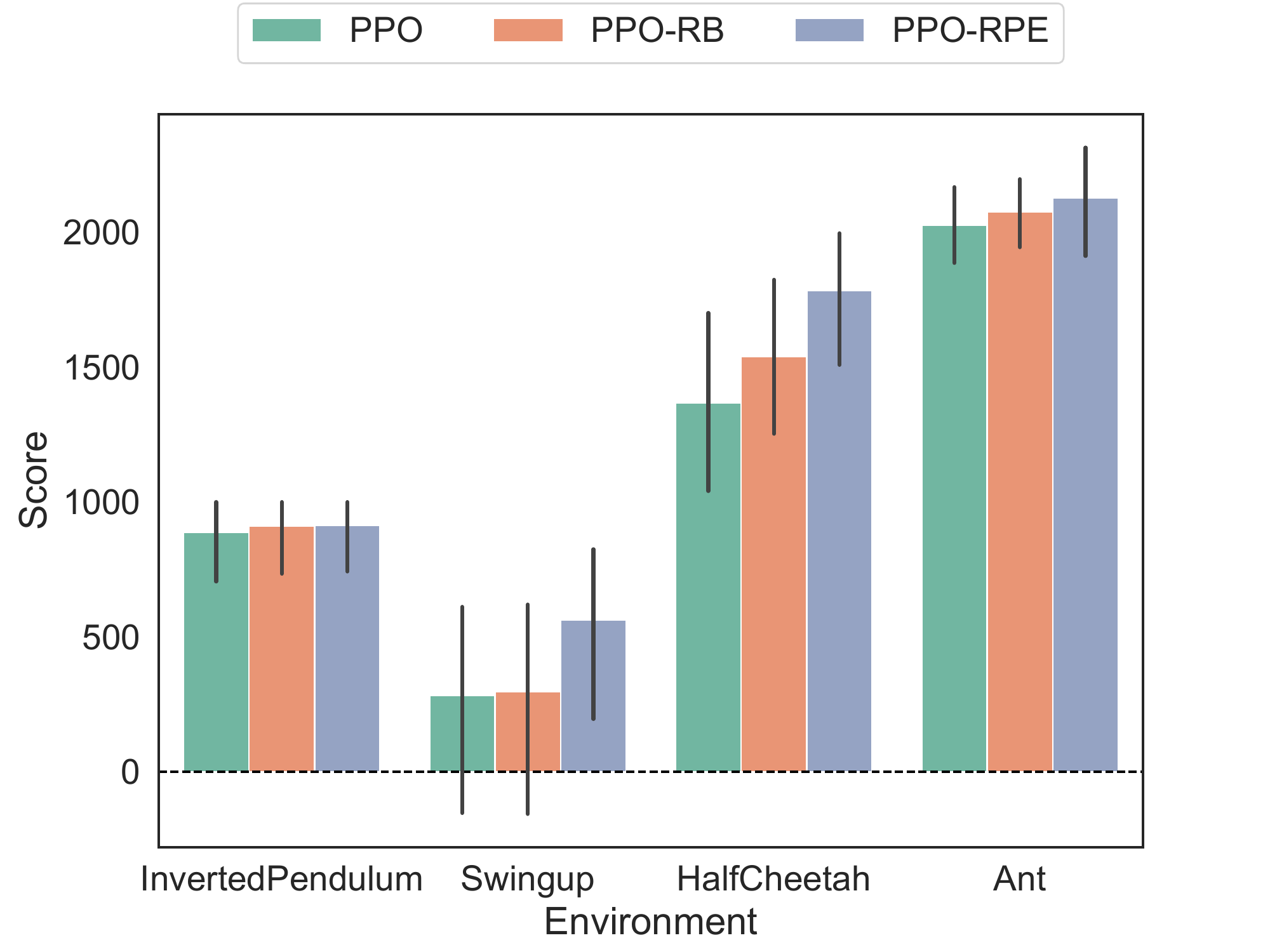}
    \caption{Summary of four benchmark tasks as bar plots:
        after learning under each condition with different random seeds, the learned policy was tested 50 times on the same environment to evaluate the median of the test scores;
        in all the tasks, PPO-RPE outperformed PPO and PPO-RB in terms of mean of the task performances.
    }
    \label{fig:sim_summary}
\end{figure}

\begin{figure*}[tb]
    \centering
    \subfigure[InvertedPendulum]{
        \includegraphics[keepaspectratio=true,width=0.23\linewidth]{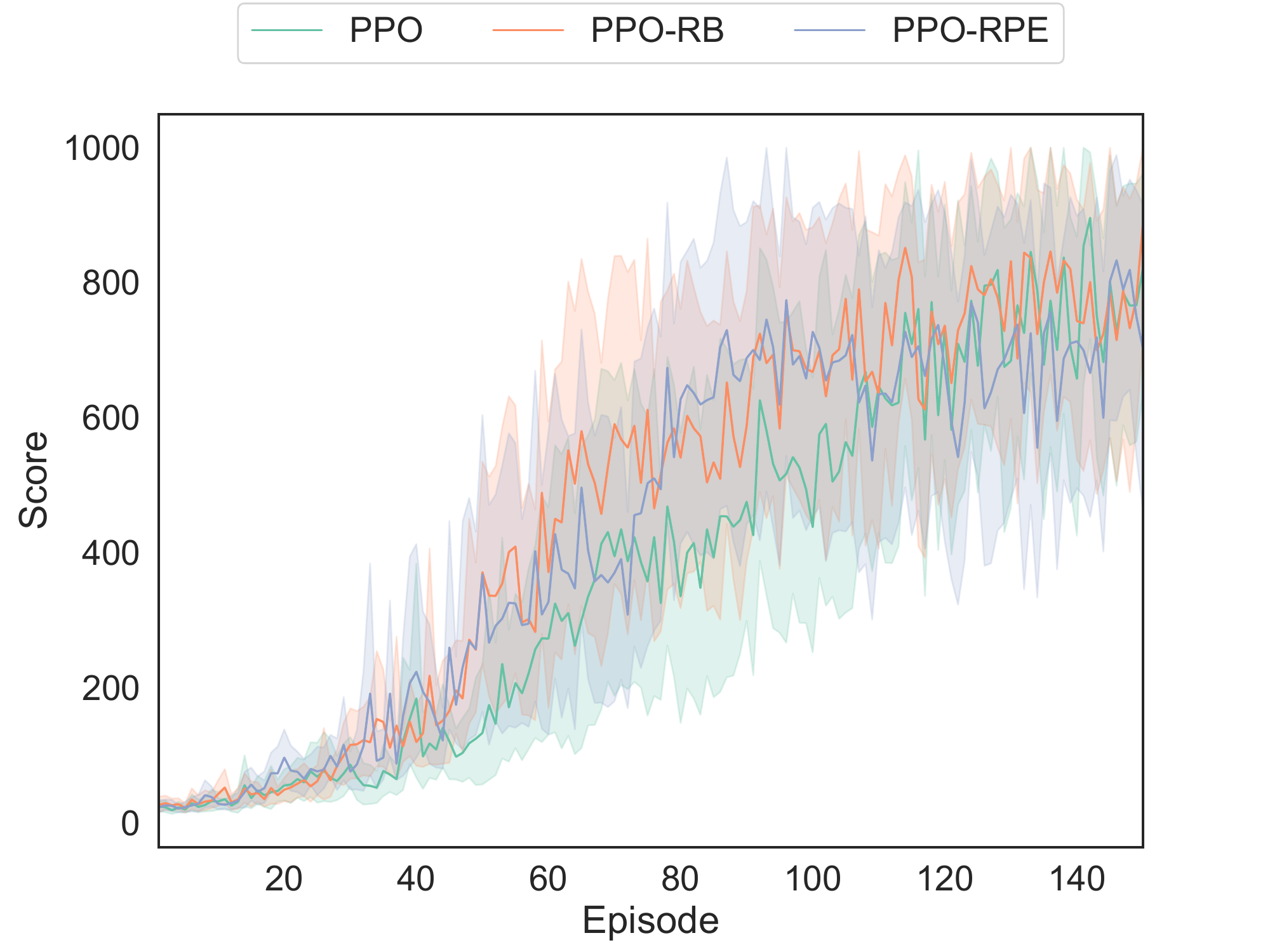}
    }
    \centering
    \subfigure[Swingup]{
        \includegraphics[keepaspectratio=true,width=0.23\linewidth]{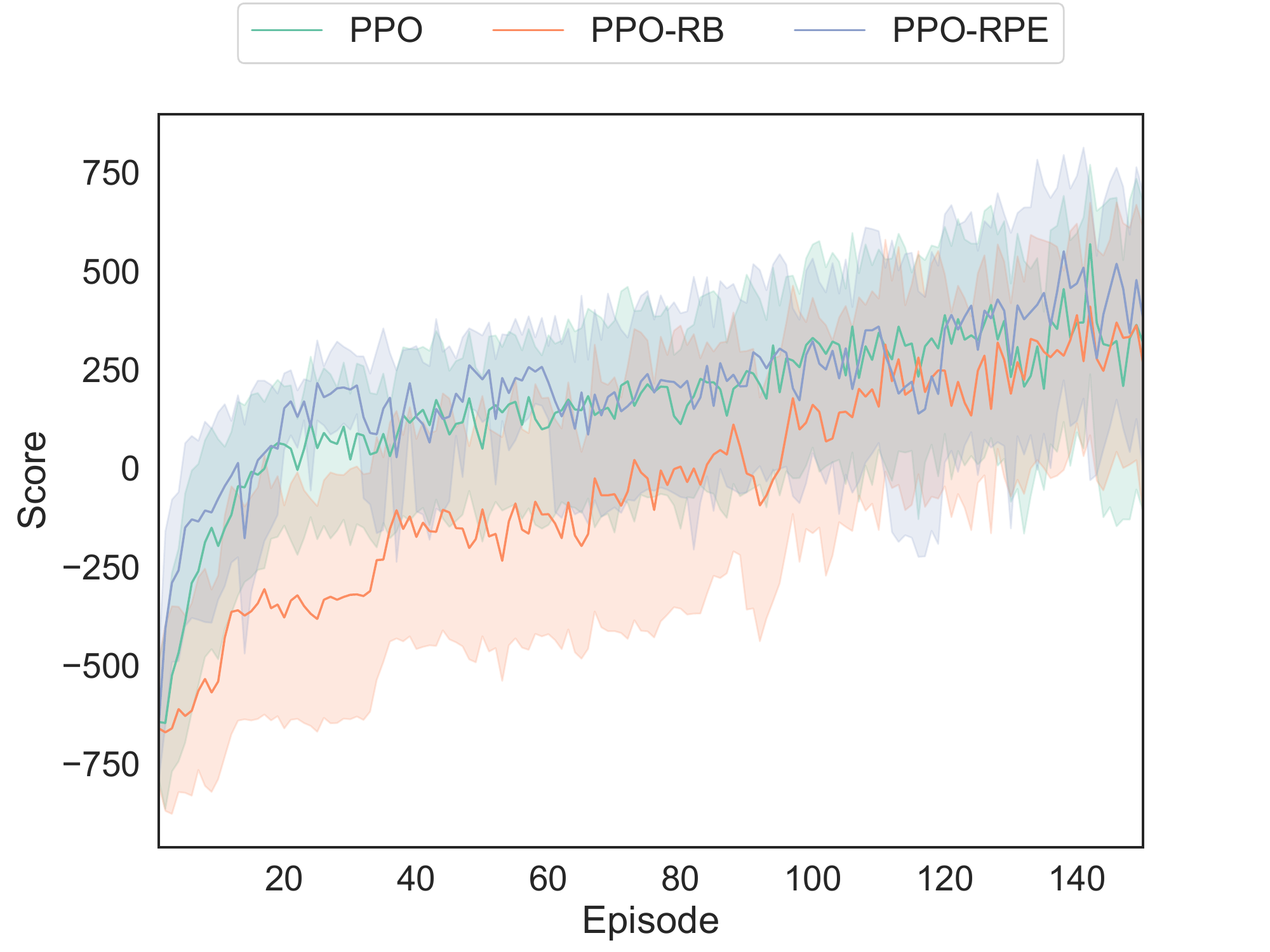}
    }
    \centering
    \subfigure[HalfCheetah]{
        \includegraphics[keepaspectratio=true,width=0.23\linewidth]{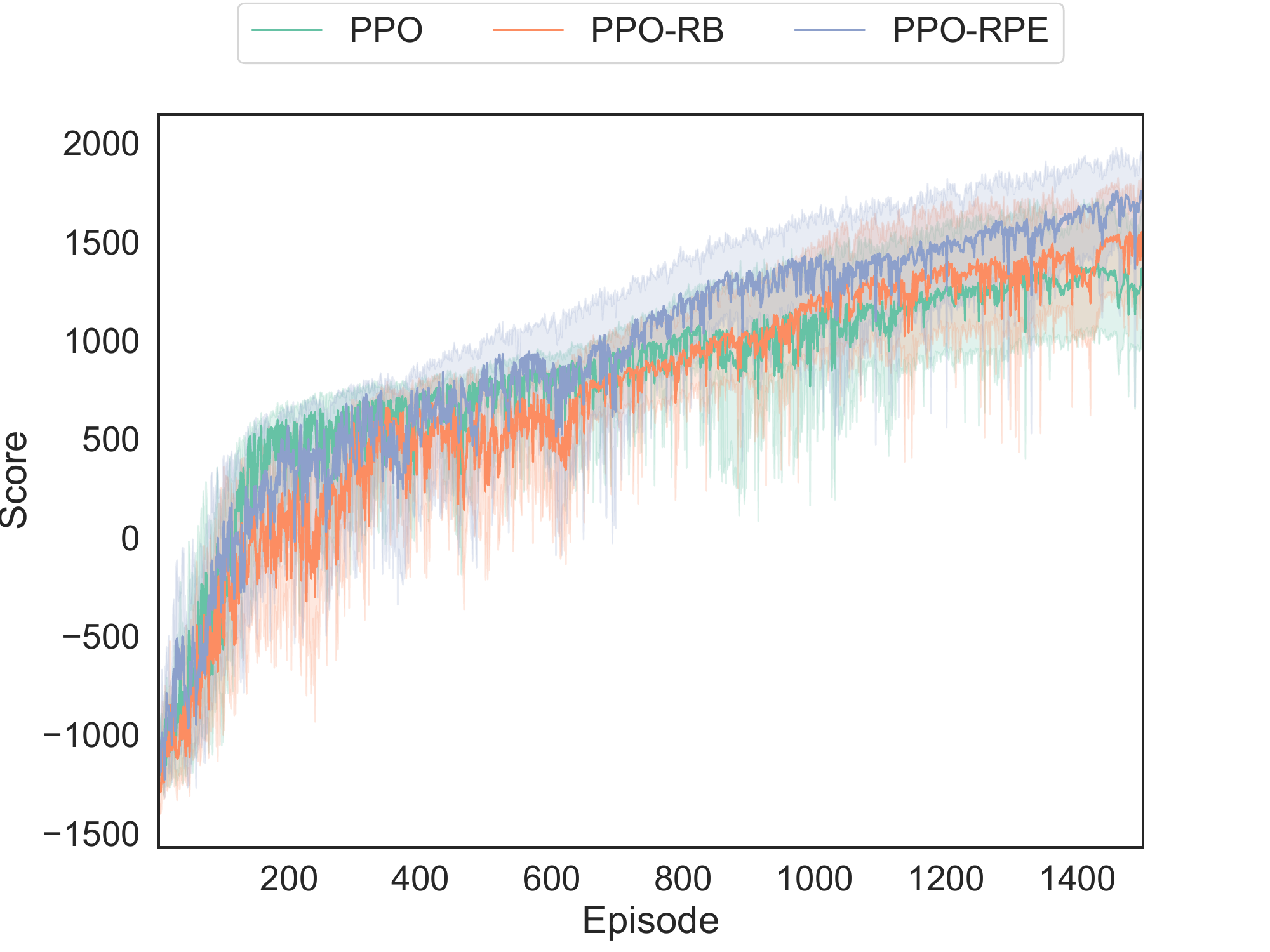}
    }
    \centering
    \subfigure[Ant]{
        \includegraphics[keepaspectratio=true,width=0.23\linewidth]{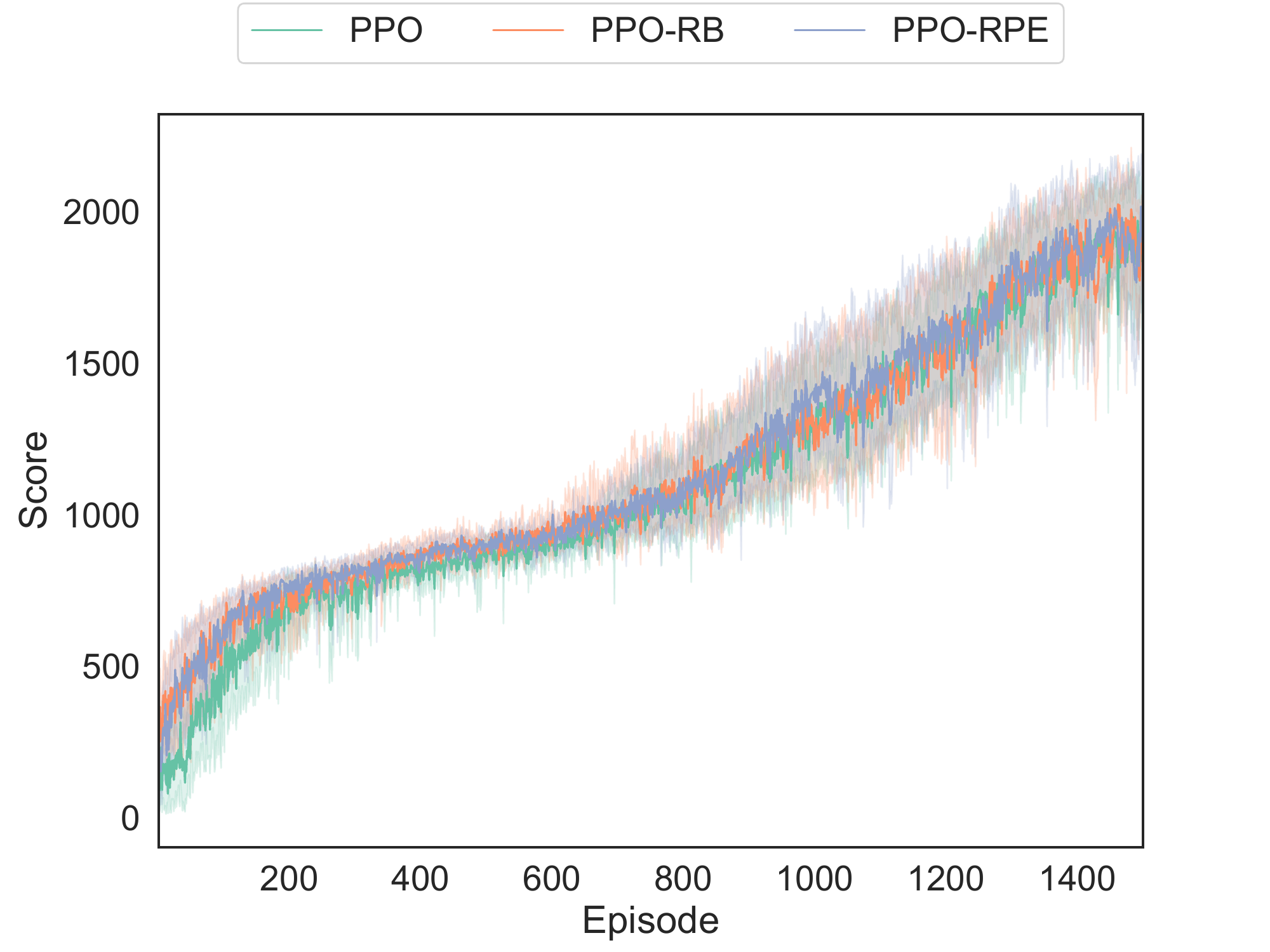}
    }
    \caption{Learning curves of four benchmark tasks:
        the sum of rewards at each episode are given as the score;
        the corresponding shaded areas show the 95~\% confidence intervals;
        only PPO-RPE succeeded in improving the score from the early stage in both (a) and (b);
        in (c), PPO-RPE was able to gradually improve the task performance over the others.
    }
    \label{fig:sim_learn}
\end{figure*}

\begin{figure*}[tb]
    \centering
    \subfigure[InvertedPendulum]{
        \includegraphics[keepaspectratio=true,width=0.23\linewidth]{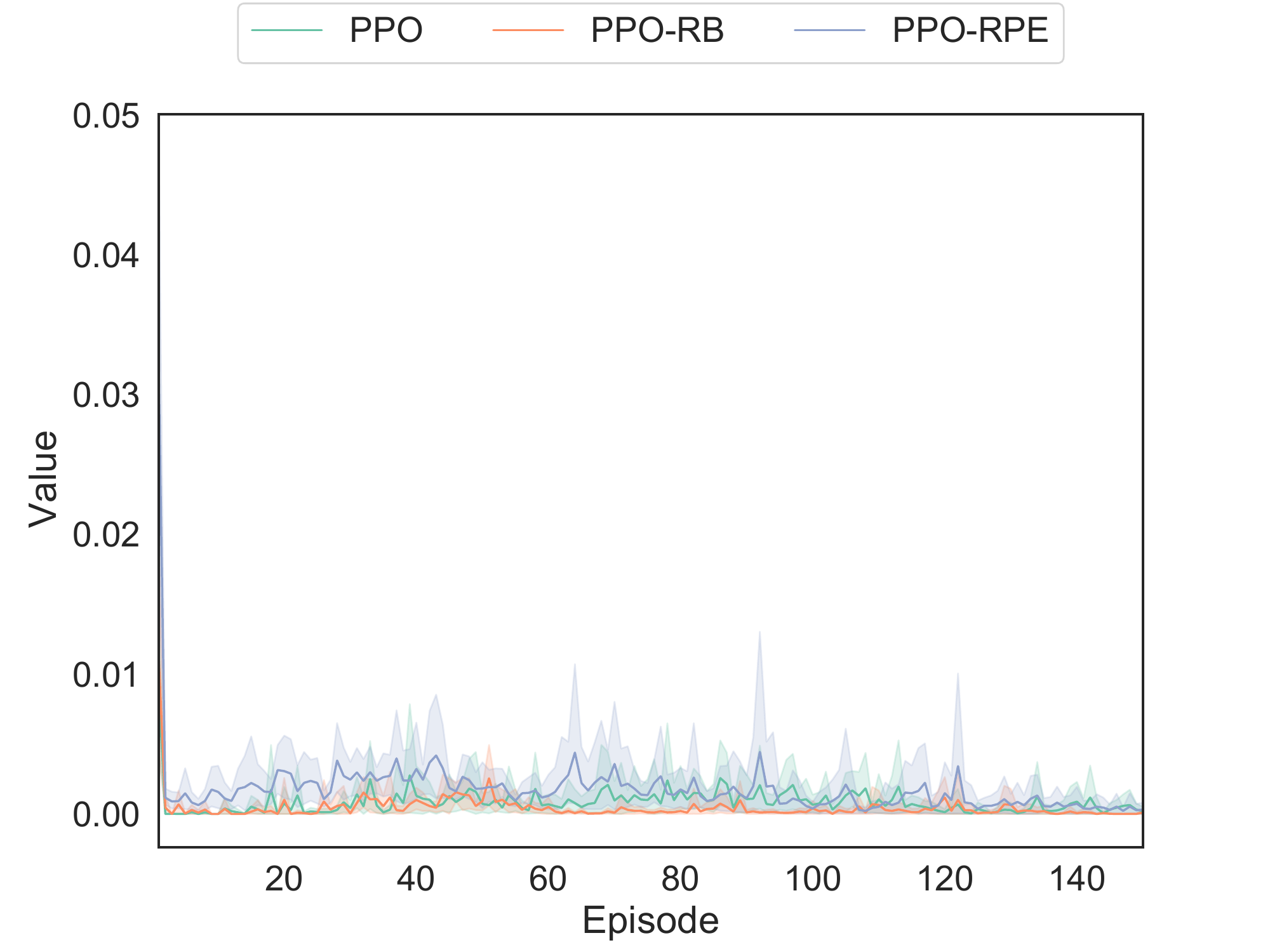}
    }
    \centering
    \subfigure[Swingup]{
        \includegraphics[keepaspectratio=true,width=0.23\linewidth]{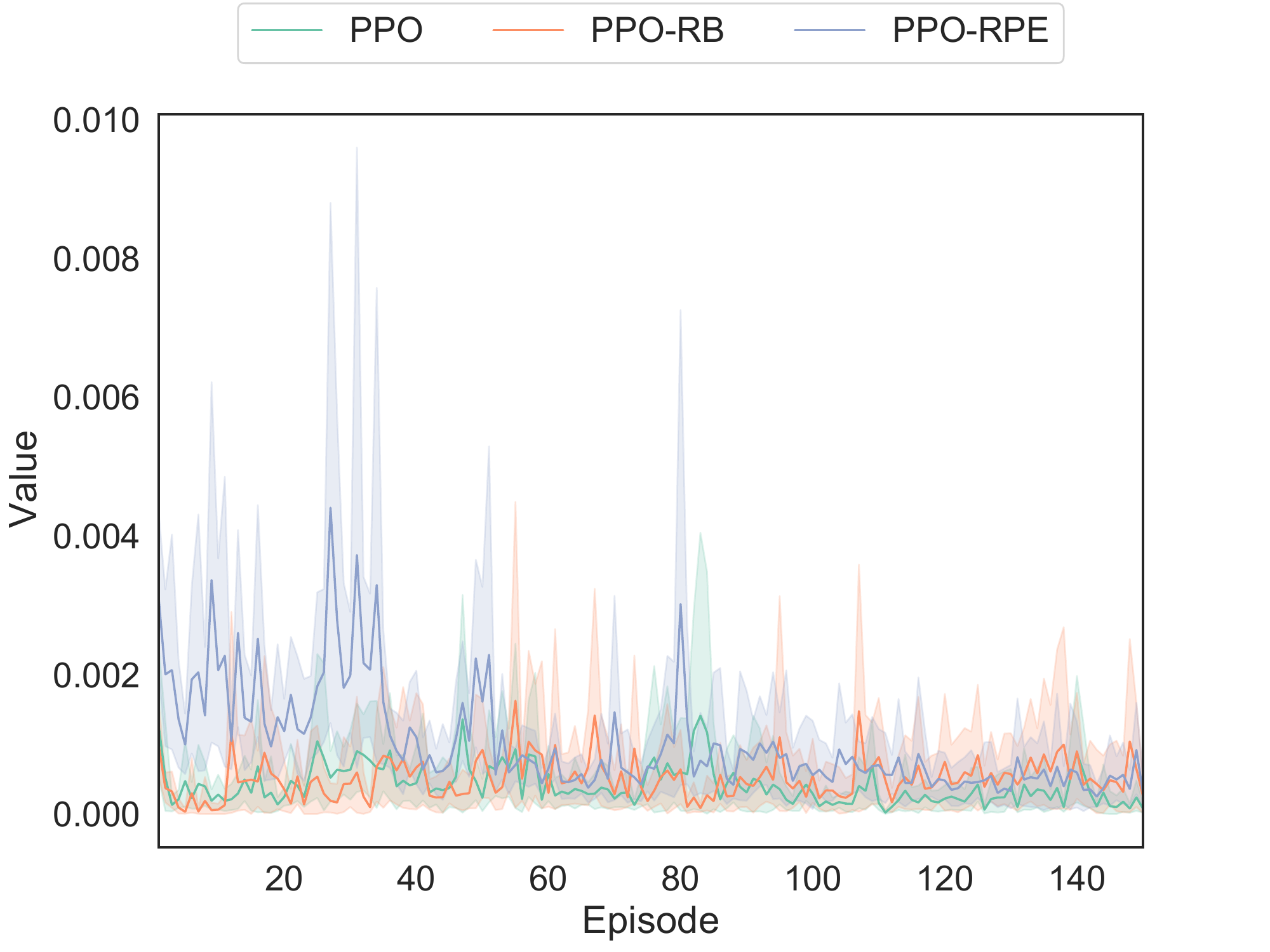}
    }
    \centering
    \subfigure[HalfCheetah]{
        \includegraphics[keepaspectratio=true,width=0.23\linewidth]{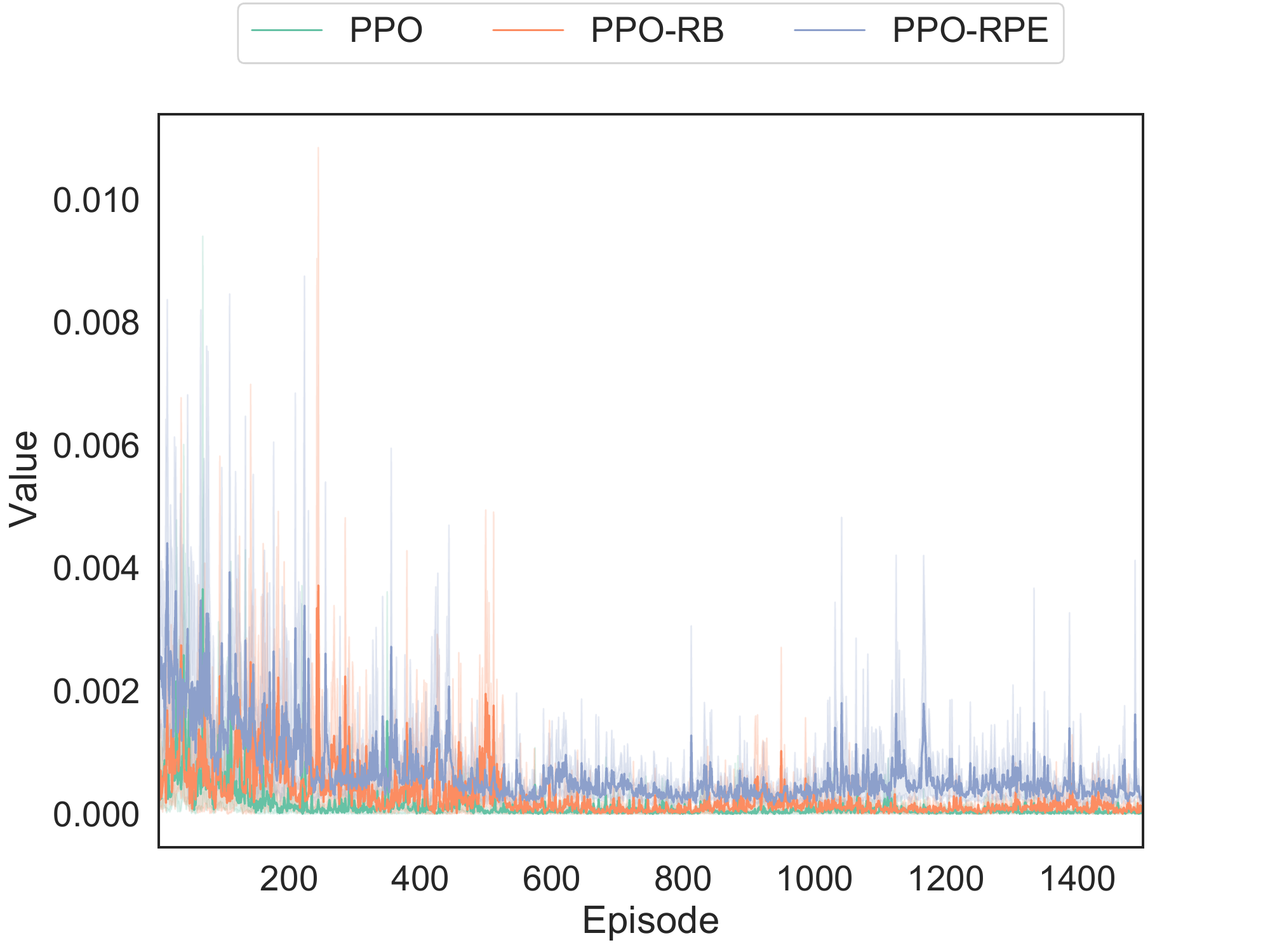}
    }
    \centering
    \subfigure[Ant]{
        \includegraphics[keepaspectratio=true,width=0.23\linewidth]{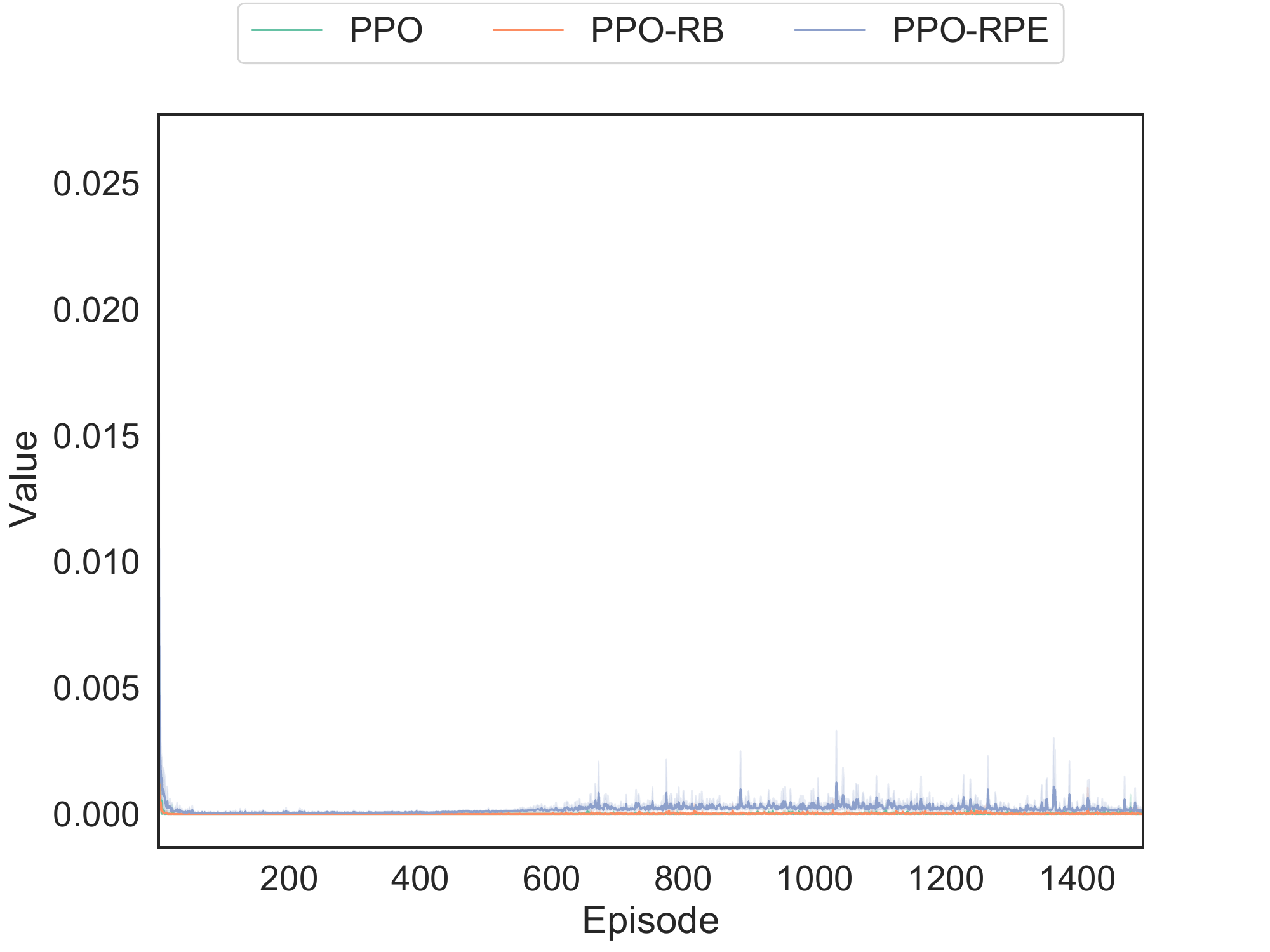}
    }
    \caption{Trajectories of the amount of regularization:
        instead of each regularization term $\Omega$, $\sigma (\rho - \rho^\dagger)$ is illustrated;
        PPO-RPE has stronger regularization than the others due to the explicit constraint between the latest and baseline policies.
    }
    \label{fig:sim_isr_diff}
\end{figure*}

\subsection{Conditions}

Four benchmark tasks are simulated by Open AI Gym with Pybullet dynamical engine~\cite{brockman2016openai,coumans2016pybullet}.
The agent performs the task using the learned policy 50 times to compute the sum of rewards for each, and their median is used as the score for each condition.
Totally, 10 trials for each condition are conducted with different random seeds.

In this paper, the policy $\pi$ is parameterized by student-t distribution~\cite{kobayashi2019student}, aiming for more robust updates of $\pi$ than one parameterized by normal distribution.
Neural networks for approximating the parameters in $\pi$ contains $L=5$ fully connected layers with $N=100$ neurons and pairs of layer normalization~\cite{ba2016layer} and Swish activation function~\cite{elfwing2018sigmoid} for nonlinearity.
For the baseline policy $b$, the policy before update, $\pi(a \mid s; \theta_\mathrm{old})$, is employed.
That is, the proposed and conventional methods add the regularization to the latest policy so that a single parameter update does not cause an extreme change in the policy.

To learn the value function, a target network with t-soft update~\cite{kobayashi2021t} is employed for stable and efficient improvement.
Learning of policy is accelerated using adaptive eligibility traces~\cite{kobayashi2020adaptive}, instead of experience replay~\cite{lin1992self} for reducing memory cost.
In addition, the policy entropy regularization based on SAC~\cite{haarnoja2018soft} with a regularization weight $\beta_{DE}$; and the TD regularization~\cite{parisi2019td} with a regularization weight $\beta_{TD}$ are combined.
As an optimizer employed in eq.~\eqref{eq:sgd}, a robust SGD, i.e., LaProp~\cite{ziyin2020laprop} with t-momentum~\cite{ilboudo2020robust} and d-AmsGrad~\cite{kobayashi2020towards} (so-called td-AmsProp), is employed with their default parameters except the learning rate.
Note that these implementations are built on PyTorch~\cite{paszke2017automatic}.

The common parameters for the above implementations are summarized in Table~\ref{tab:sim_parameter}.
Two conventional methods~\cite{schulman2017proximal,wang2020truly} and the proposed method, PPO-RPE, are compared throughout the simulations as follows:
\begin{enumerate}
    \item PPO~\cite{schulman2017proximal}: $\eta = 0$ for no rollback
    \item PPO-RB~\cite{wang2020truly}: $\eta = 0.3$ as recommended value
    \item PPO-RPE (proposal): $\beta = 0.5$ for symmetry
\end{enumerate}
Note that the threshold parameter $\epsilon = 0.1$, which was chosen from the range recommended in~\cite{schulman2017proximal} (i.e. $[0.1, 0.3]$), is commonly set as listed in Table~\ref{tab:sim_parameter}.

\subsection{Results}

To evaluate the task performance by the learned policy, the test results after learning are also depicted in Fig.~\ref{fig:sim_summary}.
Learning curves gained from 10 trials for each condition are also illustrated in Fig.~\ref{fig:sim_learn}.
Note that since the experience replay~\cite{lin1992self} was not employed unlike the major implementations, the learning tendency is very different.
In addition, the trajectories of the amount of regularization (i.e. $\sigma (\rho - \rho^\dagger)$) are shown in Fig.~\ref{fig:sim_isr_diff}.

First of all, as can be seen in Fig.~\ref{fig:sim_summary}, PPO-RPE performed as well as or better than the other methods.
PPO and PPO-RB deteriorated learning speed in the respective tasks (see Fig.~\ref{fig:sim_learn}(a) and ~(b)).
In contrast, PPO-RPE shows its excellent learning speed in both tasks.
Even in high-dimensional locomotion tasks, PPO-RPE is also effective, and its regularization effect is especially remarkable to improve the task performance in HalfCheetah task (see Fig.~\ref{fig:sim_learn}(c)), which requires conservative learning to avoid local optima.

In Fig.~\ref{fig:sim_isr_diff}, we can find that PPO-RPE has stronger constraint from the latest to baseline policies than the others.
This is because PPO-RPE always regularizes the policy except for $\pi = b$, while PPO (and PPO-RB) does not if the policy is within the threshold.
Nevertheless, it is noted that all PPO variants serve as a safeguard against extreme cases, as imagined from the small mean of $\sigma (\rho - \rho^\dagger)$ during training.

\section{Conclusion}

This paper proposed PPO-RPE, which is a variant of PPO integrated with RPE divergence regularization.
In the standard PPO, the regularization target is not explicitly given, inducing undesired behaviors like no capability to bind policy with its baseline.
Although the variant of PPO, named PPO-RB, provides the ad-hoc solution for this problem, it still remains another issue about inconsistency between symmetric threshold and asymmetric domain of density ratio, which would make regularization unbalanced.
To resolve both of them in a mathematically rigorous way, the minimization problem with RPE divergence, gain of which was designed based on the symmetric threshold for the relative density ratio, was considered as PPO-RPE.
The threshold-based gain as well as the standard PPO yields the regularization adaptive to the scale of the main minimization target (i.e. the negative advantage function).
As a result, PPO-RPE with stronger constraint to the baseline outperformed the conventional methods by enhancing learning stability.

Although the threshold allows us to design the gain of regularization intuitively, this may still have different optimal values depending on the average divergence between the latest and baseline policies.
In the near future, therefore, the divergence between their marginal policies will be investigated to design the threshold for further improvement.
As another perspective of PPO-RPE for robotic applications, the ability to appropriately bind the policy updates is expected to facilitate the management of the behaviors during the learning process.
By taking advantage of this benefit, application research of the proposed method will be promoted.

\bibliographystyle{IEEEtran}
{
\bibliography{pporpe}
}

\end{document}